\documentclass[runningheads]{llncs}
\usepackage[T1]{fontenc}
\usepackage{amsfonts}
\usepackage{multicol}
\usepackage{multirow}
\usepackage{booktabs}
\usepackage{array}
\usepackage{arydshln}
\usepackage{xcolor}
\usepackage{colortbl}
\usepackage{algorithmicx}
\usepackage{algorithm}
\usepackage[noend]{algpseudocode}
\usepackage{amsmath}
\usepackage{amssymb}
\usepackage{tabularx}
\usepackage{nicematrix}
\usepackage{subcaption}
\usepackage[backend=biber,
sorting=none,
doi=false,
isbn=false,
url=true,
maxcitenames=5,
mincitenames=1,
style=ieee,]{biblatex}
\usepackage{graphicx,verbatim}
\usepackage[misc]{ifsym}

\newcolumntype{I}{%
    @{\hspace{2.5pt}}%
    !{\vrule width 0.35pt}%
    @{\hspace{2.5pt}}%
}

\newcommand{\res}[2]{\ensuremath{#1\,(\pm #2)}} 
\newcommand{\bestres}[2]{\ensuremath{\boldsymbol{#1\,(\pm #2)}}} 
\newcommand{\secondres}[2]{\underline{\ensuremath{#1\,(\pm #2)}}}

\begin{document}
\newcolumntype{P}[1]{>{\centering\arraybackslash}p{#1}}
\titlerunning{ZeroSlide}
\title{Continual Model Merging with Test-Time Adaptation for Whole-Slide Image Analysis}
%
\begin{comment}  %% Removed for anonymized MICCAI 2025 submission
\author{First Author\inst{1}\orcidID{0000-1111-2222-3333} \and
Second Author\inst{2,3}\orcidID{1111-2222-3333-4444} \and
Third Author\inst{3}\orcidID{2222--3333-4444-5555}}
%
\authorrunning{F. Author et al.}
\institute{Princeton University, Princeton NJ 08544, USA \and
Springer Heidelberg, Tiergartenstr. 17, 69121 Heidelberg, Germany
\email{lncs@springer.com}\\
\url{http://www.springer.com/gp/computer-science/lncs} \and
ABC Institute, Rupert-Karls-University Heidelberg, Heidelberg, Germany\\
\email{\{abc,lncs\}@uni-heidelberg.de}}

\end{comment}

\author{Duc-Thanh Le$^{1,2}$, Doanh C. Bui$^{1,2}$, Maï K. Nguyen$^{3}$, Khang Nguyen$^{1,2}$}  %% Added for anonymized MICCAI 2025 submission
\authorrunning{Le et al.}
\institute{$^1$University of Information Technology \\ $^2$Viet Nam National University Ho Chi Minh City \\ $^3$ETIS (UMR 8051), CY Cergy Paris University, ENSEA, CNRS, France}

\maketitle     % typeset the header of the contribution
\begin{abstract}

Model merging offers a practical alternative to conventional continual learning by integrating independently fine-tuned models without retaining previous training data. Recent \textit{state-of-the-art model merging methods employ test-time adaptation (TTA-guided merging)} to address distribution shifts by adjusting merging-related variables using unlabeled target data. However, these methods have primarily been studied in multi-task or single-target settings, and their behavior under sequential continual learning remains insufficiently understood. We present a benchmark study that maps this family of methods to rehearsal-free continual Whole Slide Image classification and evaluates them against traditional continual-learning approaches. Experiments on six TCGA cancer-subtyping cohorts cover CLASS-IL and TASK-IL scenarios, in-domain and out-of-domain evaluation, and different task orders. The results show that adapting model merging at test time can provide strong task-specific performance and improve retention of previously acquired knowledge without storing historical WSIs. Nevertheless, performance remains sensitive to task order and to the interaction between adaptation on the current distribution and accumulated knowledge. This benchmark identifies model merging with test-time adaptation as a promising direction for continual computational pathology and motivates future methods that balance adaptation to domain shift with explicit preservation of historical knowledge.

% Authors must provide keywords and are not allowed to remove this Keyword section.

\end{abstract}

\section{Introduction}

Cancer classification from Whole Slide Images (WSIs) must accommodate continuously emerging cohorts and diagnostic tasks, while the gigapixel scale of WSIs and institutional data restrictions make centralised retraining and long-term storage of previous patient data impractical~\cite{lu2021data,shen2022federated}. This motivates continual learning methods that incorporate new knowledge without revisiting historical slides, patches, or feature bags.

Sequential fine-tuning is susceptible to catastrophic forgetting, whereas rehearsal-based methods require storing previous samples~\cite{boschini2022transfer,buzzega2020dark}. Model merging offers an exemplar-free alternative by consolidating independently fine-tuned task models directly in parameter space~\cite{ilharco2022editing,yadav2023ties}. MergeSlide~\cite{bui2026mergeslide} demonstrates the effectiveness of continual model merging for WSI classification. However, its merging rule is fixed once a new task model is incorporated and is not explicitly adapted to the incoming test distribution, potentially limiting its flexibility under distribution shift.

Test-time adaptive merging provides a natural extension to this static paradigm. AdaMerging~\cite{yang2024adamerging} adapts task-wise or layer-wise merging coefficients, AdaRank~\cite{lee2025adarank} learns masks over singular components of task vectors, and Hi-Vec~\cite{ambekar2026hierarchical} adapts hierarchical prediction layers and propagates target-specific updates through task-vector-based weight merging. Despite their different formulations, these methods follow a common pattern of adapting merging-related variables from unlabeled test samples before constructing the final inference weights. Crucially, they were originally developed for multi-task merging or standard test-time adaptation rather than continual learning. When only the current task distribution is available, such adaptation may bias the merged model toward recent data, interfere with previously accumulated knowledge, and introduce repeated update to merge to inference overhead.

In this work, we systematically transfer AdaMerging, AdaRank, and Hi-Vec to continual WSI classification over six TCGA cancer-subtyping cohorts. We compare them with exemplar-free continual learning baselines including EWC~\cite{kirkpatrick2017overcoming} and LwF~\cite{li2017learning}, and DER++~\cite{buzzega2020dark} as a rehearsal-based reference. Our study examines whether test-time adaptive merging improves current-task adaptation without increasing catastrophic forgetting, and whether repeatedly updating and reconstructing merging variables provides sufficient gains to justify its computational cost.

% \section{Related Works}

\section{Methodology}
\subsection{Preliminaries}

\paragraph{\textbf{Problem formulation.}}
Given a sequential stream $\{\mathcal{D}_t\}_{t=1}^{T}$ of $T$ cancer-subtyping tasks, where each $\mathcal{D}_t$ contains WSIs from $c_t$ subtypes, the goal is to fine-tune a slide aggregator $f_A$ on each $\mathcal{D}_t$ and incrementally merge the resulting weights into a unified model
$\tilde{\theta}_{1:t}=\mathcal{M}(\theta_0, \theta_1, \ldots, \theta_t),$
where $\theta_0$ is the shared pre-trained backbone. The task vector encoding task-specific knowledge is defined as $\tau_t = \theta_t - \theta_0.$ Two constraints are strictly enforced: (i) no access to $\mathcal{D}_{1:t-1}$ after task $t-1$ completes, and (ii) no exemplar memory. We evaluate under CLASS-IL (task identity unknown at inference) and TASK-IL (task identity provided).

\paragraph{\textbf{WSI processing.}}
Inspired by MergeSlide~\cite{bui2026mergeslide}, each WSI $X_i \in \mathcal{D}_t$ is first preprocessed via the CLAM pipeline~\cite{lu2021data}, which segments tissue regions using Otsu thresholding and tiles them into patches. Each patch $x_{i,j}$ is then passed through the frozen vision encoder of TITAN~\cite{ding2025multimodal}, a state-of-the-art pathology VLM, to extract a patch embedding$\mathbf{v}_{i,j} \in \mathbb{R}^{d_{\mathrm{vis}}}.$
We obtain a bag of patch embeddings$\mathcal{B}_i=\{\mathbf{v}_{i,j}\}_{j=1}^{|\mathcal{B}_i|}.$
Following MergeSlide, $K$ patches are randomly sampled using an index set $\mathcal{I}_i$ satisfying $|\mathcal{I}_i|=K$, forming $\mathcal{B}'_i=\{\mathbf{v}_{i,j} \mid j \in \mathcal{I}_i\} \subset \mathcal{B}_i.$
The sampled bag is then passed through the MLP-free slide aggregator $f_A(\cdot;\theta_t)$ to obtain the slide-level embedding$\mathbf{z}_i=f_A(\mathcal{B}'_i;\theta_t).$ No patch, slide, or feature bag from previous tasks is retained.

\subsection{Continual Model Merging with Test-Time Adaptation}

\begin{figure}[t]
    \centering

    % Traditional Continual Learning
    \begin{subfigure}[t]{0.90\linewidth}
        \centering
        \includegraphics[width=\linewidth]{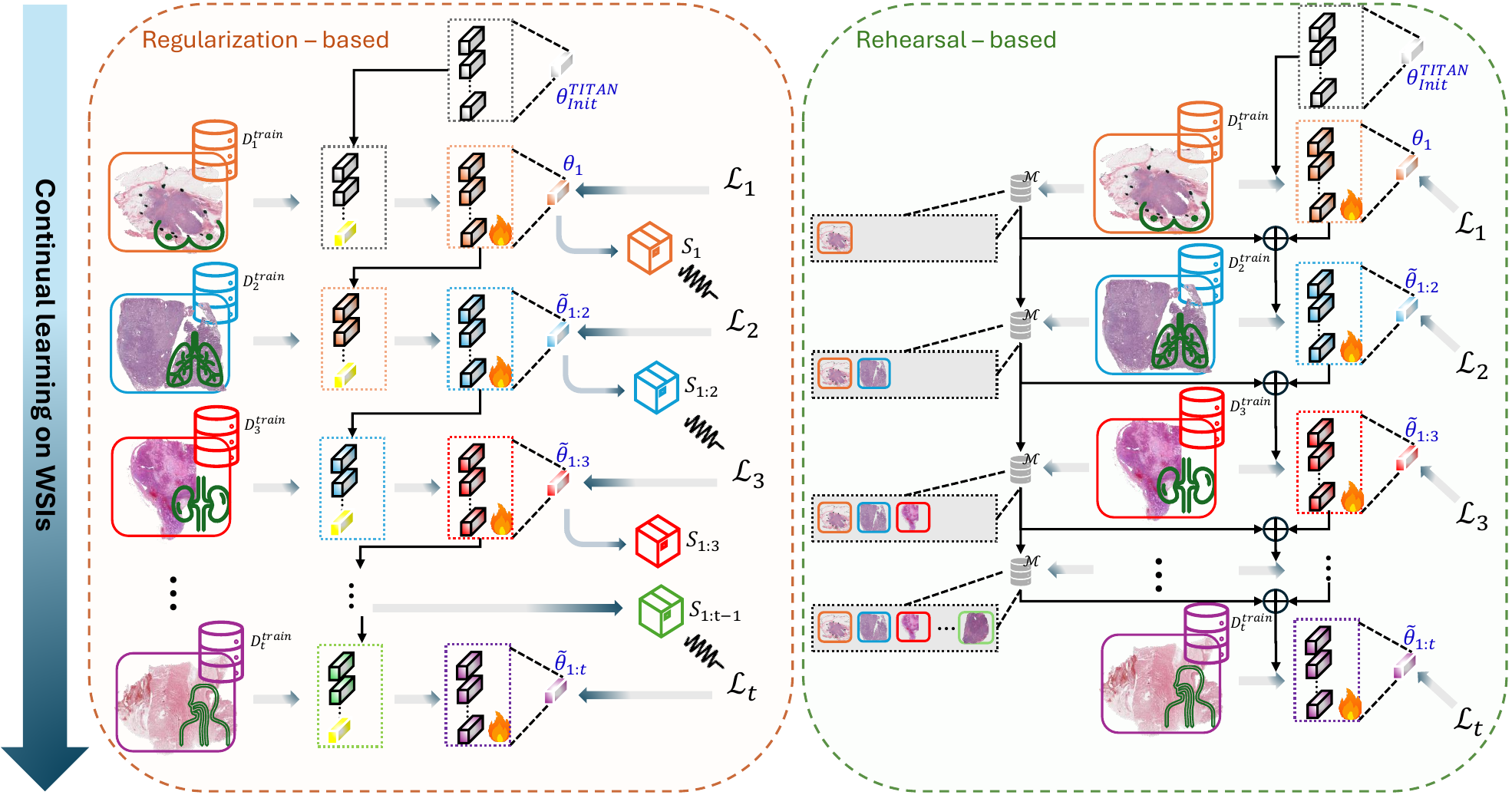}
        \label{fig:traditional_cl}
    \end{subfigure}

    \vspace{-0.4cm}

    % TTA-Guided Merging
    \begin{subfigure}[t]{0.90\linewidth}
        \centering
        \includegraphics[width=\linewidth]{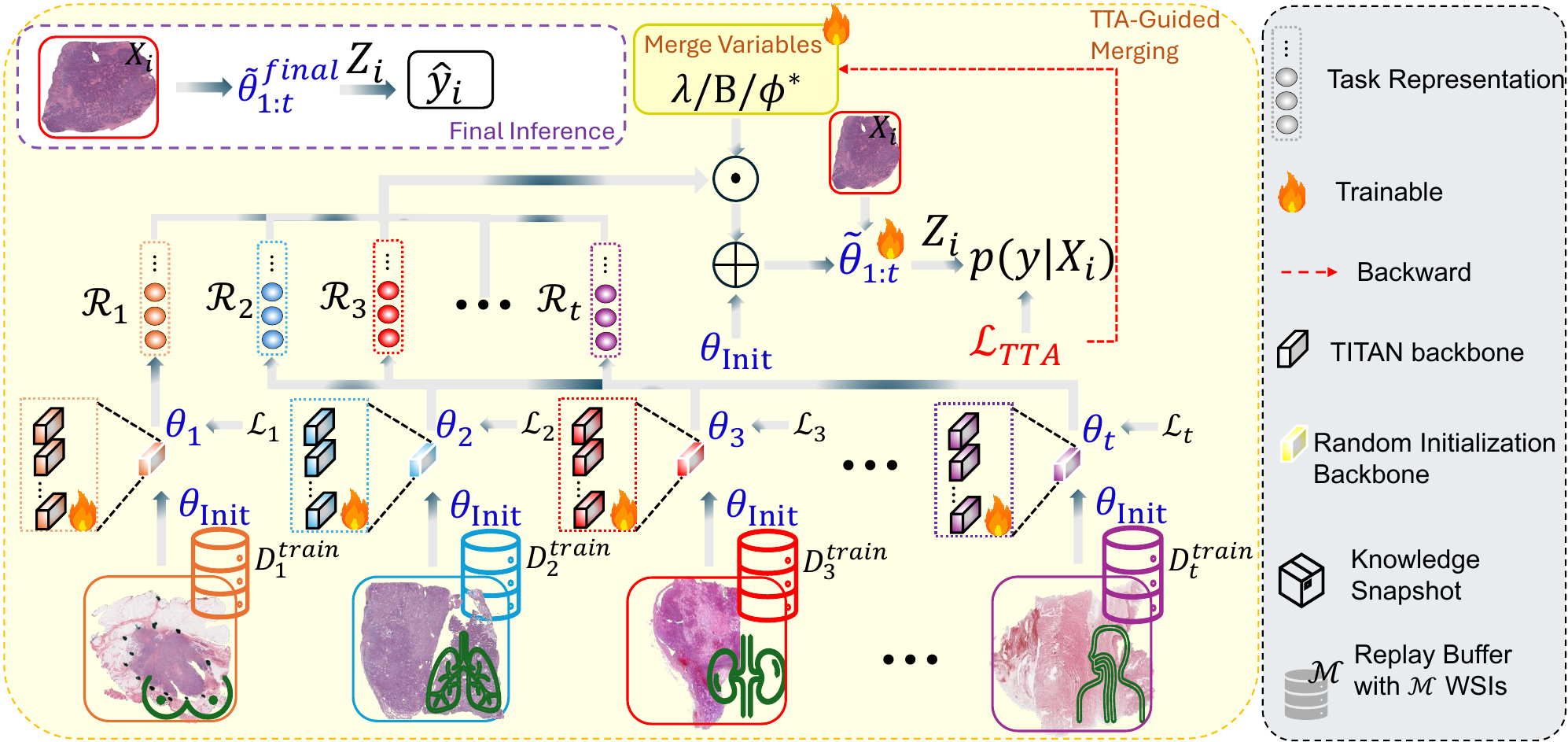}
        \label{fig:tta_guided_merging}
    \end{subfigure}

    \caption{{Comparison of traditional continual learning and TTA-guided model merging for WSIs.}}

    \label{fig:comparison}
\end{figure}

As illustrated in Fig.~\ref{fig:comparison}, each task is first fine-tuned independently from the same initialization $\theta_0$, producing a task-specific model $\theta_k$. Each method then converts this model into a method-dependent task representation $\mathcal{R}_k$: AdaMerging~\cite{yang2024adamerging} uses task vectors, AdaRank~\cite{lee2025adarank} uses their singular-component decomposition, and Hi-Vec~\cite{ambekar2026hierarchical} uses hierarchical prediction layers with task-vector-based updates. At continual step $t$, the available representations $\mathcal{R}_{1:t}$ are combined with method-specific merge variables, namely coefficients $\boldsymbol{\lambda}_t$, binary masks $\mathbf{B}_t$, or a selected hierarchical layer $\phi_t^{*}$, to construct the current model $\tilde{\theta}_{1:t}= \mathcal{M}\!\left(\theta_0,\mathcal{R}_{1:t};\boldsymbol{\xi}_t\right).$ The current unlabeled test stream $\mathcal{X}_t^{\mathrm{test}}$ is then used to optimize $\boldsymbol{\xi}_t$ through the TTA objective, after which the model is reconstructed or updated for final inference. Since the original methods are not continual, we adapt them to a rehearsal-free sequential protocol in which no previous WSIs, patches, or feature bags are retained.

\paragraph{\textbf{AdaMerging.}}
AdaMerging learns task-wise coefficients $\lambda_{k,t}$, or layer-wise coefficients $\lambda_{k,t}^{(\ell)}$, by minimizing prediction entropy on the current test stream. The task-wise merged model is
\begin{equation}
\tilde{\theta}_{1:t}(\boldsymbol{\lambda}_t)
=
\theta_0+\sum_{k=1}^{t}\lambda_{k,t}\tau_k,
\qquad
\tau_k=\theta_k-\theta_0 .
\label{eq:adamerging}
\end{equation}
For layer-wise merging, $\tilde{\theta}_{1:t}^{(\ell)}=\theta_0^{(\ell)}+\sum_{k=1}^{t}\lambda_{k,t}^{(\ell)}\tau_k^{(\ell)}$. The coefficients are optimized as $\boldsymbol{\lambda}_t^{*}=\arg\min_{\boldsymbol{\lambda}_t}\mathbb{E}_{X\sim\mathcal{X}_t^{\mathrm{test}}}[H(p_A(\cdot\mid X;\tilde{\theta}_{1:t}(\boldsymbol{\lambda}_t)))]$.

\paragraph{\textbf{AdaRank.}}
AdaRank replaces fixed top-$k$ SVD truncation with binary masks that select useful singular components. For task $k$ and layer $\ell$, let $\tau_k^{(\ell)}=U_k^{(\ell)}\Sigma_k^{(\ell)}V_k^{(\ell)\top}$ and $B_{k,t}^{(\ell)}\in\{0,1\}^{1\times m_\ell}$. The merged layer is
\begin{equation}
\tilde{\theta}_{1:t}^{(\ell)}(\mathbf{B}_t^{(\ell)})
=
\theta_0^{(\ell)}
+
\lambda_t^{(\ell)}
\sum_{k=1}^{t}
U_k^{(\ell)}
\left(
\operatorname{diag}(B_{k,t}^{(\ell)})
\odot
\Sigma_k^{(\ell)}
\right)
V_k^{(\ell)\top}.
\label{eq:adarank}
\end{equation}
The masks are optimized by entropy minimization on $\mathcal{X}_t^{\mathrm{test}}$, using the Straight-Through Estimator to retain binary values in the forward pass while enabling gradient-based optimization.

\paragraph{\textbf{Hi-Vec.}}
Hi-Vec replaces a single classification head with hierarchical linear layers $\Phi=\{\phi_1,\ldots,\phi_R\}$ of increasing dimensionality. For each test batch, dynamic layer selection chooses $\phi_t^{*}=\arg\min_{\phi\in\Phi}\|\nabla_{W_\phi}\mathcal{L}_{\mathrm{TTA},t}\|_2$. Hierarchical agreement is measured by $I(p_{\phi_t^{*}};p_\phi)=H(p_{\phi_t^{*}})-H(p_{\phi_t^{*}}\mid p_\phi)$; adaptation is suppressed when the average agreement is below $\delta_{\mathrm{OOD}}$. Finally, target information is shared with layers satisfying $\operatorname{cos}(\mathbf{v}_{\phi_t^{*},t},\mathbf{v}_{\phi,t})>\delta_{\mathrm{sim}}$, where $\mathbf{v}_{\phi,t}=\operatorname{vec}(W_{\phi,t})$, through $W_{\phi,t}[\phi_t^{*}]\leftarrow W_{\phi_t^{*},t}+\alpha W_{\phi,t}[\phi_t^{*}]$.

\paragraph{\textbf{Test-Time Adaptation Objective.}}
Despite their distinct architectural mechanisms, AdaMerging, AdaRank, and Hi-Vec share a common surrogate objective for unsupervised test-time adaptation: Shannon entropy minimization over the unlabeled test stream. Given a WSI $X \in \mathcal{X}_t^{\mathrm{test}}$ and the predictive distribution $p_A(y=c \mid X;\theta)$ over $C$ classes, the TTA objective at continual step $t$ is defined as
\begin{equation}
\mathcal{L}_{\mathrm{TTA},t}(\theta)
=
\mathbb{E}_{X \sim \mathcal{X}_t^{\mathrm{test}}}
\left[
-\sum_{c=1}^{C}
p_A(y=c \mid X;\theta)
\log p_A(y=c \mid X;\theta)
\right].
\label{eq:tta_entropy_loss}
\end{equation}
Although the objective is shared, each method applies $\mathcal{L}_{\mathrm{TTA},t}$ to a different set of adaptation variables:
\begin{itemize}
    \item \textbf{AdaMerging} optimizes the task-wise or layer-wise merging coefficients by computing $\nabla_{\boldsymbol{\lambda}_t}\mathcal{L}_{\mathrm{TTA},t}$.

    \item \textbf{AdaRank} backpropagates $\mathcal{L}_{\mathrm{TTA},t}$ through the continuous proxy variables associated with the binary masks $\mathbf{B}_t$, using the Straight-Through Estimator to select singular components.

    \item \textbf{Hi-Vec} first uses the layer-wise gradient norm $\|\nabla_{W_\phi}\mathcal{L}_{\mathrm{TTA},t}\|_2$ for dynamic layer selection, and then minimizes $\mathcal{L}_{\mathrm{TTA},t}$ with respect to the selected layer $\phi_t^{*}$ and the shared encoder parameters.
\end{itemize}

\section{Experiment}

\paragraph{\textbf{Datasets.}}
We evaluate on six cancer subtyping cohorts sourced from the Genomic Data Commons (GDC) portal of The Cancer Genome Atlas (TCGA)~\footnote{https://portal.gdc.cancer.gov/}: TCGA-BRCA (breast), TCGA-RCC (kidney), TCGA-NSCLC (lung), TCGA-ESCA (esophagus), TCGA-TGCT (testis), and TCGA-CESC (cervix uteri). Each cohort is partitioned into 10 folds; 10-fold cross-validation is applied uniformly across all methods. Dataset statistics are summarized in Tab.~\ref{tab:wsi_datasets}

\begin{table}[h]
    \centering
    \caption{Dataset statistics for the six TCGA cohorts. The cohorts span two groups by data availability: \textit{common cohorts} (BRCA, RCC, NSCLC) with abundant WSIs, and \textit{rare cohorts} (ESCA, TGCT, CESC) with limited slide counts and pronounced class imbalance, posing distinct challenges for both per-task fine-tuning and TTA-guided merging coefficient optimization.}
    \label{tab:wsi_datasets}
    \scriptsize
    \setlength{\tabcolsep}{2pt}
    \renewcommand{\arraystretch}{1.0}

    \begin{tabularx}{\columnwidth}{
        @{}
        c
        >{\raggedright\arraybackslash}p{2.4cm}
        >{\raggedright\arraybackslash}X
        r
        c
        @{}
    }
        \toprule
        \textbf{Task} &
        \textbf{Dataset} &
        \textbf{Subtype} &
        \textbf{\#WSIs} &
        \textbf{Ratio} \\
        \midrule

        \multirow{2}{*}{1}
        & \multirow{2}{=}{TCGA-BRCA (B)}
        & invasive ductal carcinoma (IDC)
        & 726 & 4.9 \\
        & & invasive lobular carcinoma (ILC)
        & 149 & 1 \\

        \midrule
        \multirow{3}{*}{2}
        & \multirow{3}{=}{TCGA-RCC (R)}
        & clear cell renal cell carcinoma (CC)
        & 498 & 4.2 \\
        & & papillary renal cell carcinoma (P)
        & 289 & 2.4 \\
        & & chromophobe renal cell carcinoma (ChRCC)
        & 118 & 1 \\

        \midrule
        \multirow{2}{*}{3}
        & \multirow{2}{=}{TCGA-NSCLC (N)}
        & squamous cell carcinoma (SCC)
        & 845 & 7.8 \\
        & & adenocarcinoma (A)
        & 109 & 1 \\

        \midrule
        \multirow{2}{*}{4}
        & \multirow{2}{=}{TCGA-ESCA (E)}
        & squamous cell carcinoma (SCC)
        & 114 & 1.3 \\
        & & adenocarcinoma (A)
        & 86 & 1 \\

        \midrule
        \multirow{2}{*}{5}
        & \multirow{2}{=}{TCGA-TGCT (T)}
        & seminoma (S)
        & 66 & 2.3 \\
        & & mixed germ cell tumor (MGCT)
        & 29 & 1 \\

        \midrule
        \multirow{2}{*}{6}
        & \multirow{2}{=}{TCGA-CESC (C)}
        & squamous cell carcinoma (SCC)
        & 270 & 5.5 \\
        & & adenocarcinoma (A)
        & 49 & 1 \\

        \bottomrule
    \end{tabularx}
\end{table}

% \paragraph{\textbf{Implementation Details.}}
% All WSIs are tiled into non-overlapping \(256\times256\) patches at \(10\times\) magnification. Patch embeddings are extracted using the frozen vision encoder of TITAN, yielding 768-dimensional representations. All methods use the same Transformer-based slide aggregator \(f_{\mathcal{A}}\), initialized from TITAN's pretrained weights \(\theta_{\mathrm{base}}\), together with a task-specific linear classification head for each task. At inference, CLASS-IL predictions are obtained by taking the argmax over the concatenated output space of all task heads, whereas TASK-IL uses only the head corresponding to the designated task. To ensure a controlled comparison, all methods follow the same patch-sampling protocol during training. Specifically, each slide is represented by (K) randomly sampled patch bags, following the per-task fine-tuning procedure of MergeSlide~\cite{bui2026mergeslide}. This protocol is applied consistently to TTA-guided merging, regularization-based, and rehearsal-based methods; the methods differ only in their continual-learning objective, model-merging strategy, and, where applicable, the use of test-time adaptation at task boundaries. All models are trained for \(N_e=10\) epochs per task using AdamW~\cite{loshchilov2017decoupled}. Rehearsal-based methods are evaluated with buffers containing 10 or 30 WSIs. All experiments are conducted on a single NVIDIA A100-SXM4 GPU with 80 GB of memory.

\paragraph{\textbf{Implementation Details.}}
WSIs are partitioned into non-overlapping \(256\times256\) patches at \(10\times\) magnification, and 768-dimensional patch embeddings are extracted using the frozen TITAN vision encoder. All methods share the same TITAN-initialized Transformer slide aggregator \(f_{\mathcal{A}}\) and task-specific linear heads. CLASS-IL inference is performed over the concatenated outputs of all task heads, whereas TASK-IL uses only the head associated with the given task. Following MergeSlide~\cite{bui2026mergeslide}, the continual-learning and TTA-guided merging methods represent each slide using \(K\) randomly sampled patch embeddings under a common sampling protocol. In contrast, naive fine-tuning and fully supervised training use all available patch embeddings during both training and inference. Models are trained for \(N_e=10\) epochs per task using AdamW~\cite{loshchilov2017decoupled}. DER++ is evaluated with replay buffers of 10 and 30 WSIs. All experiments are conducted on a single NVIDIA A100-SXM4 GPU with 80\,GB of memory.

\paragraph{\textbf{Evaluation Metrics.}}
We report bACC for CLASS-IL, where task identity is unavailable, and masked bACC for TASK-IL, where the task label is provided. Both use mean per-class recall to account for class imbalance. We further report Macro F1, which assigns equal importance to all cancer subtypes, and Weighted F1, which weights each class according to its prevalence. These complementary metrics assess minority-class recognition and overall performance under the observed class distribution. Final mean accuracy (Mean ACC), forgetting (FGT), and backward transfer (BWT) are also reported. CLASS-IL bACC remains the primary metric because it represents the more challenging deployment setting.

\paragraph{\textbf{In/Out-of-Domain Evaluation.}}
We evaluate all methods under in-domain (IND) and out-of-domain (OOD) protocols. IND permits site overlap between training and test sets, whereas OOD restricts test slides to institutions unseen during training~\cite{cheng2021robust}. The OOD setting therefore assesses whether TTA-guided merging can adapt to site-induced distribution shifts using only unlabeled target data.

% \paragraph{\textbf{Main Results.}}
% Under the original task sequence, TTA-guided merging performs strongly in TASK-IL. On in-domain (IND) results, as showed in Tab.~\ref{tab:ind_results}, Hi-Vec and AdaRank achieve masked bACC values of $94.6061\%$ and $92.8223\%$, exceeding the fully supervised reference by $3.7748$ and $1.9991$ percentage points, respectively. Under out-of-domain (OOD) shift, as showed in Tab.~\ref{tab:ood_results_brnetc}, Hi-Vec remains the strongest method with $94.1407\%$, outperforming fully supervised training by $7.311$ points.

% In CLASS-IL, AdaMerging and Hi-Vec outperform the strongest continual-learning baseline in bACC by $7.5124$ and $7.8192$ percentage points on IND, and by $5.6321$ and $5.9499$ points on OOD, respectively. Although their mean accuracy remains below DER++ with a buffer of 30 WSIs, both methods provide substantially better knowledge retention. On IND data, AdaMerging and Hi-Vec achieve FGT values of only $2.2073\%$ and $2.6654\%$, with corresponding BWT values of $-1.4375\%$ and $-2.3088\%$, compared with $26.8298\%$ FGT and $-26.4299\%$ BWT for DER++. A similar trend is observed under OOD shift, where their FGT values remain low at $2.2404\%$ and $2.6285\%$, with BWT values of $-1.9990\%$ and $-2.2167\%$, respectively. In contrast, AdaRank achieves competitive TASK-IL performance but exhibits greater forgetting in CLASS-IL.

% =========================================================
% Combined IND and OOD results: B -> R -> N -> E -> T -> C
% =========================================================
\begin{table}[h]
\centering
\caption{\textbf{Comparison of TTA-guided merging and conventional
continual learning methods under the IND and OOD settings on six TCGA
datasets with the task sequence
B$\rightarrow$R$\rightarrow$N$\rightarrow$E$\rightarrow$T$\rightarrow$C}.
Results are reported as mean $\pm$ standard deviation in $\%$.
Within each setting, \textbf{bold} and \underline{underlined} values
denote the best and second-best results, respectively.}
\label{tab:main_results}
\scriptsize
\setlength{\tabcolsep}{1.8pt}
\renewcommand{\arraystretch}{1.05}
\setlength{\aboverulesep}{0.45ex}
\setlength{\belowrulesep}{0.45ex}

\resizebox{\columnwidth}{!}{%
\begin{tabular}{c I l I c I c c c I c c}
\toprule

\multirow[c]{2}{*}{\textbf{Setting}} &
\multirow[c]{2}{*}{\textbf{Method}} &
\multirow[c]{2}{*}{\textbf{Buffer}} &
\textbf{bACC $\uparrow$} &
\textbf{Masked bACC $\uparrow$} &
\textbf{Mean ACC $\uparrow$} &
\multirow[c]{2}{*}{\textbf{FGT $\downarrow$}} &
\multirow[c]{2}{*}{\textbf{BWT $\uparrow$}} \\
& & &
\textbf{(CLASS-IL)} &
\textbf{(TASK-IL)} &
\textbf{(CLASS-IL)} &
& \\

\midrule

% ========================= IND =========================
\multirow{9}{*}{\textbf{IND}}
& Naive Finetuning & --
& \res{67.1044}{3.2976}
& \res{90.5364}{2.0347}
& \res{89.1803}{1.2534}
& \res{32.1120}{4.2359}
& \res{-31.6681}{4.3679} \\

& Fully Supervised & --
& \bestres{85.0110}{2.3417}
& \res{90.8313}{1.8376}
& --
& --
& -- \\

\cmidrule{2-8}

& LwF~\cite{li2017learning} & --
& \res{65.1804}{4.4238}
& \res{90.5335}{1.8113}
& \res{88.7447}{1.7930}
& \res{33.1937}{5.7010}
& \res{-33.1457}{5.6773} \\

& EWC~\cite{kirkpatrick2017overcoming} & --
& \res{64.4093}{2.4041}
& \res{88.9731}{2.1901}
& \res{87.5168}{1.4701}
& \res{34.7054}{2.6241}
& \res{-34.4042}{2.7596} \\

& DER++~\cite{buzzega2020dark} & 10 WSIs
& \res{69.2483}{4.4018}
& \res{89.6904}{2.2743}
& \secondres{89.9486}{1.6597}
& \res{27.7758}{4.9862}
& \res{-27.2555}{5.0663} \\

& DER++~\cite{buzzega2020dark} & 30 WSIs
& \res{70.8106}{3.7026}
& \res{90.0952}{2.3719}
& \bestres{90.6375}{1.5345}
& \res{26.8298}{4.7636}
& \res{-26.4299}{4.6563} \\

\cmidrule{2-8}

& AdaMerging~\cite{yang2024adamerging} & --
& \res{78.3230}{2.0489}
& \res{89.8991}{1.8120}
& \res{86.9496}{1.8086}
& \bestres{2.2073}{0.7298}
& \bestres{-1.4375}{0.9317} \\

& AdaRank~\cite{lee2025adarank} & --
& \res{67.1714}{3.0644}
& \secondres{92.8223}{2.3648}
& \res{86.6967}{1.6308}
& \res{22.1014}{3.7348}
& \res{-21.8679}{3.7708} \\

& Hi-Vec~\cite{ambekar2026hierarchical} & --
& \secondres{78.6298}{2.1341}
& \bestres{94.6061}{0.7345}
& \res{87.4145}{1.8542}
& \secondres{2.6654}{0.9230}
& \secondres{-2.3088}{1.1964} \\

\midrule

% ========================= OOD =========================
\multirow{9}{*}{\textbf{OOD}}
& Naive Finetuning & --
& \res{62.9423}{4.7075}
& \res{86.2913}{3.1959}
& \res{88.8731}{2.1782}
& \res{32.6840}{6.6490}
& \res{-32.3411}{6.8363} \\

& Fully Supervised & --
& \bestres{79.3905}{5.0927}
& \res{86.8297}{3.2221}
& --
& --
& -- \\

\cmidrule{2-8}

& LwF~\cite{li2017learning} & --
& \res{63.4133}{5.8392}
& \res{86.8093}{2.6748}
& \res{87.5284}{2.5118}
& \res{31.5872}{7.9211}
& \res{-31.5276}{7.9176} \\

& EWC~\cite{kirkpatrick2017overcoming} & --
& \res{60.8898}{4.0825}
& \res{84.8140}{3.1537}
& \res{87.4747}{2.0428}
& \res{34.2521}{5.4305}
& \res{-34.0708}{5.5881} \\

& DER++~\cite{buzzega2020dark} & 10 WSIs
& \res{65.7675}{5.8657}
& \res{87.1824}{3.2936}
& \secondres{90.0000}{1.6328}
& \res{30.2324}{7.6581}
& \res{-29.8480}{7.7467} \\

& DER++~\cite{buzzega2020dark} & 30 WSIs
& \res{71.2885}{4.0662}
& \res{88.1483}{3.1624}
& \bestres{90.5644}{1.3354}
& \res{21.6273}{3.3084}
& \res{-21.2918}{3.1684} \\

\cmidrule{2-8}

& AdaMerging~\cite{yang2024adamerging} & --
& \res{76.9206}{1.4339}
& \secondres{88.5119}{1.6805}
& \res{88.6469}{0.5660}
& \bestres{2.2404}{0.7536}
& \bestres{-1.9990}{0.8746} \\

& AdaRank~\cite{lee2025adarank} & --
& \res{67.9055}{5.6059}
& \res{86.7569}{3.6485}
& \res{87.1590}{1.4127}
& \res{16.2504}{3.9810}
& \res{-15.8924}{4.3479} \\

& Hi-Vec~\cite{ambekar2026hierarchical} & --
& \secondres{77.2384}{1.5898}
& \bestres{94.1407}{0.8933}
& \res{88.9163}{0.6548}
& \secondres{2.6285}{0.6480}
& \secondres{-2.2167}{0.9649} \\

\bottomrule
\end{tabular}%
}
\end{table}

\paragraph{\textbf{Main Results.}}
As shown in Tabs.~\ref{tab:main_results}, TTA-guided merging is particularly effective in TASK-IL. Hi-Vec exceeds the fully supervised reference by \(3.77 \%\) on IND and \(7.31 \%\) on OOD, while AdaRank provides a \(-1.99\%\) gain on IND. In CLASS-IL, AdaMerging and Hi-Vec outperform the strongest continual-learning baseline in bACC by \(7.51\%\) and \(7.82\%\)  on IND, and by \(5.63\%\) and \(5.95\%\) on OOD, respectively. Although DER++ with 30 replay WSIs attains higher Mean ACC, AdaMerging and Hi-Vec preserve previous tasks more effectively: relative to DER++, they reduce FGT by approximately \(24.2\%\)--\(24.6\%\) and improve BWT by \(24.1\%\)--\(25.0\%\) on IND, with corresponding gains of about \(19\%\) under OOD shift. AdaRank remains competitive in TASK-IL but exhibits greater forgetting and stronger task dependence in CLASS-IL.

% \paragraph{\textbf{Sensitivity to task order}.}
% Under the reversed sequence, as shown in Tab.~\ref{tab:ind_results_reverse}, TTA-guided merging retains strong TASK-IL performance. In CLASS-IL, AdaMerging and Hi-Vec remain competitive without storing previous WSIs, reaching bACC values of $78.6677\%$ and $78.6292\%$. Both methods also exhibit substantially lower forgetting than naive fine-tuning, EWC, LwF, and DER++ with 10 replay WSIs, achieving FGT values of $14.3921\%$ and $14.6877\%$ and corresponding BWT values of $-13.9168\%$ and $-14.1682\%$. Although these results indicate some sensitivity to task order, they show that test-time adaptive merging continues to preserve task-specific knowledge and provides a favorable stability--plasticity trade-off under a substantially different learning sequence.
\begin{table}[h]
\centering
\caption{\textbf{Comparison of TTA-guided merging methods mapped to the
continual learning setting against conventional continual learning
methods on a benchmark of six TCGA datasets in the IND setting under
the reversed task sequence
E$\rightarrow$T$\rightarrow$C$\rightarrow$B$\rightarrow$R$\rightarrow$N}.
All results are reported as mean $\pm$ standard deviation in $\%$. \textbf{Bold} denotes the best result, while
\underline{underlined} denotes the second-best result.}
\label{tab:ind_results_reverse}
\scriptsize
\setlength{\tabcolsep}{2pt}
\renewcommand{\arraystretch}{1.05}

% Leave gaps between vertical and horizontal rules.
\setlength{\aboverulesep}{0.45ex}
\setlength{\belowrulesep}{0.45ex}

\resizebox{\columnwidth}{!}{%
\begin{tabular}{l I c I c c c I c c}
\toprule

\multirow[c]{2}{*}{\textbf{Method}} &
\multirow[c]{2}{*}{\textbf{Buffer}} &
\textbf{bACC $\uparrow$} &
\textbf{Masked bACC $\uparrow$} &
\textbf{Mean ACC $\uparrow$} &
\multirow[c]{2}{*}{\textbf{FGT $\downarrow$}} &
\multirow[c]{2}{*}{\textbf{\#BWT $\uparrow$}} \\
& &
\textbf{(CLASS-IL)} &
\textbf{(TASK-IL)} &
\textbf{(CLASS-IL)} &
& \\

\midrule

Naive Finetuning & --
& \res{61.9742}{4.1518}
& \res{90.5923}{1.6786}
& \res{61.5699}{2.2538}
& \res{41.3195}{5.7577}
& \res{-41.1791}{5.8597} \\

Fully Supervised & --
& \bestres{85.0261}{3.4855}
& \res{90.2535}{2.4831}
& --
& --
& -- \\

\midrule

LwF~\cite{li2017learning} & --
& \res{55.5284}{4.1972}
& \res{90.3635}{1.8555}
& \res{61.7891}{1.4478}
& \res{48.6558}{4.1489}
& \res{-48.4978}{4.2601} \\

EWC~\cite{kirkpatrick2017overcoming} & --
& \res{60.4054}{4.9796}
& \res{90.6885}{1.3657}
& \res{61.3409}{2.0593}
& \res{42.4943}{6.5684}
& \res{-42.3889}{6.6177} \\

DER++~\cite{buzzega2020dark} & 10 WSIs
& \res{73.8973}{5.7857}
& \res{90.6087}{1.5475}
& \secondres{80.0842}{2.9199}
& \res{21.9437}{8.1397}
& \res{-21.7424}{8.1899} \\

DER++~\cite{buzzega2020dark} & 30 WSIs
& \secondres{80.3150}{4.8132}
& \res{91.1414}{1.3826}
& \bestres{86.3063}{1.8397}
& \bestres{11.8343}{5.6468}
& \bestres{-11.5218}{5.8428} \\

\midrule

AdaMerging~\cite{yang2024adamerging} & --
& \res{78.6677}{2.5413}
& \res{90.0052}{2.1859}
& \res{75.1242}{0.9795}
& \secondres{14.3921}{0.7467}
& \secondres{-13.9168}{0.9786} \\

AdaRank~\cite{lee2025adarank} & --
& \res{65.8651}{2.8355}
& \secondres{93.5975}{1.7599}
& \res{74.3150}{1.3113}
& \res{26.7375}{3.6149}
& \res{-26.2774}{3.6550} \\

Hi-Vec~\cite{ambekar2026hierarchical} & --
& \res{78.6292}{1.8153}
& \bestres{94.2190}{1.1353}
& \res{75.1268}{1.0787}
& \res{14.6877}{0.9015}
& \res{-14.1682}{1.0023} \\

\bottomrule
\end{tabular}%
}
\end{table}

\paragraph{\textbf{Sensitivity to Task Order.}}
Under the reversed sequence, Hi-Vec and AdaRank retain strong TASK-IL performance, exceeding the fully supervised reference by \(3.97\%\) and \(3.34\%\), respectively (Tab.~\ref{tab:ind_results_reverse}). In CLASS-IL, AdaMerging and Hi-Vec outperform DER++ with 10 replay WSIs in bACC by approximately \(4.8\%\) while requiring no exemplar storage. They also reduce FGT by \(7.3\%\)--\(7.6\%\) and improve BWT by \(7.6\%\)--\(7.8\%\) relative to this rehearsal baseline. DER++ with 30 WSIs remains stronger in final accuracy and forgetting. Compared with the original task order, the higher FGT of AdaMerging and Hi-Vec indicates sensitivity to task sequence, although both continue to preserve prior knowledge more effectively than naive fine-tuning, LwF, EWC, and limited replay.

% =========================================================
% IND reversed: E -> T -> C -> B -> R -> N
% =========================================================

\begin{figure}[h]
    \centering
    \begin{subfigure}[t]{0.48\textwidth}
        \centering
        \includegraphics[width=\linewidth]
        {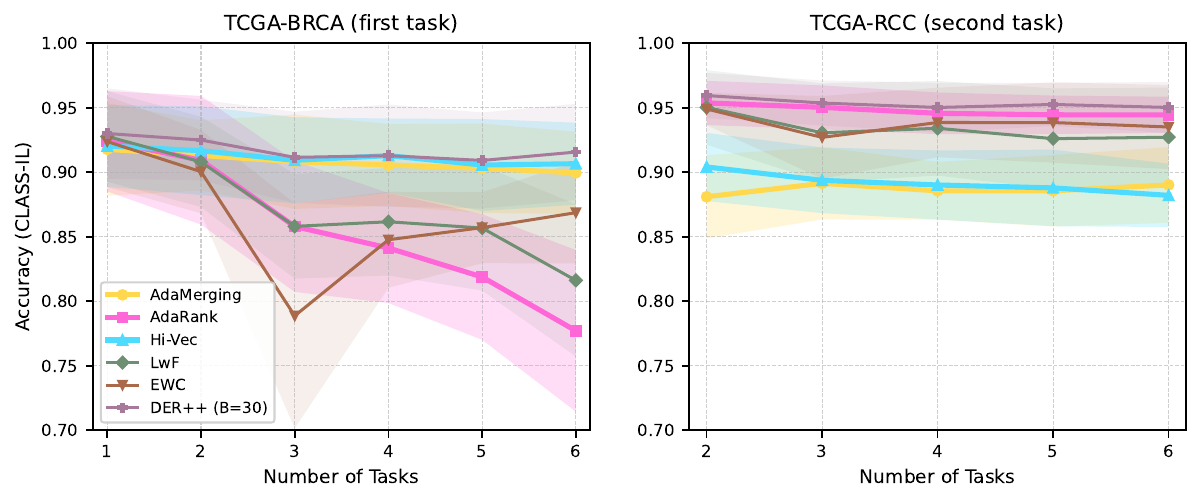}
        \caption{Performance drop under the IND setting.}
        \label{fig:performance_drop_ind}
    \end{subfigure}
    \hfill
    \begin{subfigure}[t]{0.48\textwidth}
        \centering
        \includegraphics[width=\linewidth]
        {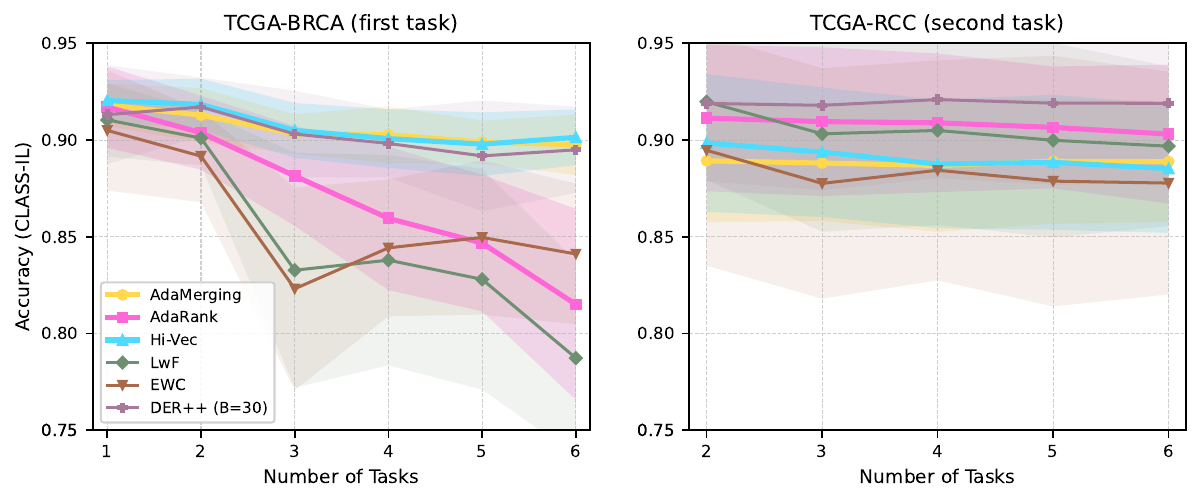}
        \caption{Performance drop under the OOD setting.}
        \label{fig:performance_drop_ood}
    \end{subfigure}

    \caption{Performance degradation of all methods under the
    (a) in-distribution (IND) and (b) out-of-distribution (OOD)
    evaluation settings.}
    \label{fig:performance_drop_all}
\end{figure}

\paragraph{\textbf{Performance retention}.}
We further compare the performance degradation of different methods on TCGA-BRCA and TCGA-RCC as additional tasks are incrementally introduced, as shown in Fig.~\ref{fig:performance_drop_all}. The performance trajectories further support the effectiveness of TTA-guided merging for preserving previously acquired knowledge. For the earliest task, TCGA-BRCA, AdaMerging and Hi-Vec maintain nearly flat accuracy curves under both IND and OOD evaluation, with only a small reduction after five additional tasks are incorporated. Their retention is comparable to DER++ with a 30-WSI replay buffer, while clearly exceeding LwF, EWC, and AdaRank, which exhibit substantially larger degradation on the same task. A similarly stable trend is observed on TCGA-RCC, although the relative ranking varies across methods. In particular, AdaRank preserves RCC well but degrades markedly on BRCA, indicating greater task-dependent sensitivity.

\paragraph{\textbf{Robustness to Class Imbalance.}}

% =========================================================
% Macro-F1 and Weighted-F1 across domain settings/task orders
% =========================================================
\begin{table*}[b]
    \centering
    \caption{
        \textbf{CLASS-IL F1 performance across domain settings and task orders.}
        Results are reported as mean $\pm$ standard deviation in $\%$.
        Buffer size denotes the number of stored WSIs.
        The best and second-best results within each setting are shown in
        \textbf{bold} and \underline{underlined}, respectively.
    }
    \label{tab:f1_results_all_settings}

    \scriptsize
    \setlength{\tabcolsep}{1.1pt}
    \renewcommand{\arraystretch}{1.08}
    \setlength{\aboverulesep}{0.4ex}
    \setlength{\belowrulesep}{0.4ex}

    \resizebox{\textwidth}{!}{%
    \begin{tabular}{@{}l c cc cc cc@{}}
        \toprule

        \multirow{2}{*}{\textbf{Method}}
        & \multirow{2}{*}{\textbf{Buffer}}
        & \multicolumn{2}{c}{
            \textbf{IND: B$\rightarrow$R$\rightarrow$N$\rightarrow$E$\rightarrow$T$\rightarrow$C}}
        & \multicolumn{2}{c}{
            \textbf{IND: E$\rightarrow$T$\rightarrow$C$\rightarrow$B$\rightarrow$R$\rightarrow$N}}
        & \multicolumn{2}{c}{
            \textbf{OOD: B$\rightarrow$R$\rightarrow$N$\rightarrow$E$\rightarrow$T$\rightarrow$C}} \\

        \cmidrule(lr){3-4}
        \cmidrule(lr){5-6}
        \cmidrule(lr){7-8}

        & 
        & \textbf{Macro F1 $\uparrow$}
        & \textbf{Weighted F1 $\uparrow$}
        & \textbf{Macro F1 $\uparrow$}
        & \textbf{Weighted F1 $\uparrow$}
        & \textbf{Macro F1 $\uparrow$}
        & \textbf{Weighted F1 $\uparrow$} \\

        \midrule

        % Lower and supervised references
        Naive Finetuning
        & 0
        & \res{65.3453}{3.6051}
        & \res{70.9762}{2.6980}
        & \res{58.6971}{4.5078}
        & \res{53.5409}{4.7451}
        & \res{59.6386}{5.8910}
        & \res{80.9950}{5.2650} \\

        Fully Supervised
        & 0
        & \bestres{85.3864}{2.1435}
        & \bestres{87.9087}{1.5202}
        & \bestres{85.1684}{3.1539}
        & \bestres{86.9111}{2.6209}
        & \bestres{77.0707}{4.0406}
        & \bestres{88.4054}{2.1422} \\

        \midrule

        % Traditional continual learning
        LwF~\cite{li2017learning}
        & 0
        & \res{63.7732}{4.8689}
        & \res{69.3601}{4.0187}
        & \res{51.0719}{4.0343}
        & \res{46.0991}{2.8564}
        & \res{59.2777}{6.2418}
        & \res{78.6101}{4.5714} \\

        EWC~\cite{kirkpatrick2017overcoming}
        & 0
        & \res{62.1551}{3.0931}
        & \res{68.6135}{1.9022}
        & \res{57.7269}{5.7170}
        & \res{52.6368}{4.9695}
        & \res{57.1155}{4.9927}
        & \res{79.1614}{4.4218} \\

        DER++~\cite{buzzega2020dark}
        & 10
        & \res{67.9337}{5.1597}
        & \res{73.1176}{3.2798}
        & \res{73.5335}{6.8056}
        & \res{74.9349}{8.1890}
        & \res{62.5181}{7.1758}
        & \res{82.8260}{5.8393} \\

        DER++~\cite{buzzega2020dark}
        & 30
        & \res{69.7959}{4.4529}
        & \secondres{74.2296}{2.8412}
        & \secondres{80.9389}{5.2564}
        & \secondres{84.2264}{4.2574}
        & \secondres{69.6086}{3.7791}
        & \secondres{85.4789}{3.8914} \\

        \midrule

        % Model merging with test-time adaptation
        AdaMerging~\cite{yang2024adamerging}
        & 0
        & \res{72.2490}{1.9463}
        & \res{71.6450}{1.4975}
        & \res{72.6805}{2.4349}
        & \res{72.2672}{1.5983}
        & \res{69.2330}{2.3672}
        & \res{82.6715}{1.2637} \\

        AdaRank~\cite{lee2025adarank}
        & 0
        & \res{66.3656}{3.4847}
        & \res{69.2786}{3.0901}
        & \res{65.0004}{2.9488}
        & \res{68.0237}{2.5309}
        & \res{63.3955}{3.9783}
        & \res{77.7542}{3.3980} \\

        Hi-Vec~\cite{ambekar2026hierarchical}
        & 0
        & \secondres{72.6111}{2.1860}
        & \res{71.8841}{1.5477}
        & \res{72.5704}{1.8272}
        & \res{71.9460}{1.3651}
        & \res{69.4768}{2.5831}
        & \res{82.7992}{1.5840} \\

        \bottomrule
    \end{tabular}%
    }
\end{table*}

As shown in Tab.~\ref{tab:f1_results_all_settings}, Macro F1 and Weighted F1 provide complementary views of performance on highly imbalanced cancer-subtyping cohorts, with the former emphasizing minority-class recognition and the latter reflecting the observed class distribution. Across the evaluated settings, the TTA-guided merging methods generally outperform the regularization-based continual-learning baselines and remain competitive with replay-based DER++. AdaMerging and Hi-Vec are particularly consistent, achieving strong Macro F1 while maintaining Weighted F1 at a similar level; DER++ with a 30-WSI buffer remains the strongest competitor in several settings. The comparatively small gap between Macro and Weighted F1 further suggests that TTA-guided merging is less dominated by majority classes and provides a more balanced treatment of common and rare cancer subtypes.

\section{Discussion}
Mapping TTA-guided merging methods to a continual stream exposes a pronounced stability--plasticity trade-off because adaptation is driven only by the current unlabeled test distribution. In AdaMerging~\cite{yang2024adamerging}, entropy minimization at step $t$ may shift coefficients associated with earlier task vectors, biasing the merged model toward the most recent task. In AdaRank~\cite{lee2025adarank}, updating binary masks without access to previous data may suppress singular components that remain important for earlier tasks. Hi-Vec~\cite{ambekar2026hierarchical} faces a related gating dilemma: a strict agreement threshold may reject genuinely novel target samples, whereas a permissive threshold can allow excessive parameter updates and representation drift. These risks arise from the absence of an explicit constraint that preserves previously acquired subspaces during adaptation. Our results, particularly the increased forgetting under the reversed task order and the stronger task dependence of AdaRank, indicate that TTA-guided merging can remain effective but is sensitive to how current-task adaptation interacts with accumulated knowledge. This motivates future mechanisms that combine test-time plasticity with explicit protection of historical parameter subspaces.

\section{Conclusion}
This benchmark study examined state-of-the-art model merging methods that use test-time adaptation to address domain shift when transferred to rehearsal-free continual WSI classification. Across six TCGA cohorts, these approaches showed strong task-specific performance and, in several settings, improved knowledge retention over traditional continual-learning methods. However, their behavior remained sensitive to task order and to adaptation driven only by the current test distribution. These findings position model merging with test-time adaptation as a promising direction for continual computational pathology and motivate future methods that improve adaptability while explicitly protecting previously acquired knowledge.
% \section*{Acknowledgments}

\section*{Disclosure of Interests} The authors declare that they have no conflict of interest.
\printbibliography
\end{document}